\documentclass[conference]{IEEEtran}
\IEEEoverridecommandlockouts

\ifdefined\pdfobjcompresslevel
\fi

\usepackage{graphicx}
\usepackage{amsmath,amssymb,mathtools}
\usepackage{array,booktabs,makecell,multirow,tabularx}
\usepackage[ruled,vlined,linesnumbered]{algorithm2e}
\SetKwInput{Input}{Input}
\SetKwInput{Output}{Output}
\usepackage{xcolor}
\usepackage{xspace}
\usepackage{cite}
\usepackage[hidelinks]{hyperref}

\newcommand{\model}{PFedQFT\xspace}

\newcolumntype{Y}{>{\centering\arraybackslash}X}

\newcommand{\algstage}[1]{\tcc{#1}}

\title{PFedQFT: Personalized Federated Quantum-Inspired Fine-Tuning for Medical Large Language Models}

\author{
\IEEEauthorblockN{
Qihan Huang\textsuperscript{1,$\dagger$},
Lingxia Zhang\textsuperscript{1,3,$\dagger$},
Guye Zhang\textsuperscript{1,3},
Zizhu Wang\textsuperscript{1,3,*},
Xiaowei Mao\textsuperscript{1,2,3,4,*,$\ddagger$}
}
\IEEEauthorblockA{
\textsuperscript{1}Institute of Fundamental and Frontier Sciences,\\
University of Electronic Science and Technology of China, Chengdu, Sichuan, China\\
\textsuperscript{2}Sichuan Provincial Key Laboratory for Human Disease Gene Study and the Center for Medical Genetics,\\
Department of Laboratory Medicine, Sichuan Academy of Medical Sciences and Sichuan Provincial People's Hospital,\\
University of Electronic Science and Technology of China, Chengdu, Sichuan, China\\
\textsuperscript{3}Key Laboratory of Quantum Physics and Photonic Quantum Information,\\
Ministry of Education, University of Electronic Science and Technology of China, Chengdu, Sichuan, China\\
\textsuperscript{4}Sichuan Quantum Science and Technology Institute, Chengdu,  Sichuan, China\\
xiaowei.mao@uestc.edu.cn
}
\thanks{\textsuperscript{$\dagger$}These authors contributed equally.}
\thanks{\textsuperscript{*}Corresponding authors.}
\thanks{\textsuperscript{$\ddagger$}Lead contact.}
\thanks{This work was supported by the National Natural Science Foundation of China (T2322002), the National Science and Technology Major Project (2024ZD0523900), and the Sichuan Science and Technology Program (2024ZYD0006, 2024NSFTD0032, and 2025ZNSFSC0003) to X.M.}
}

\begin{document}
\bstctlcite{IEEEtran:BSTcontrol}
\maketitle

\begin{abstract}
Medical data remain fragmented across institutional silos because of privacy, governance, and ownership constraints. Federated learning enables medical institutions to adapt large language models (LLMs) collaboratively without transferring raw records. However, independently parameterized low-rank adapters incur repeated parameter and communication costs across Transformer projections, while uniform aggregation may limit adaptation to heterogeneous institutional data. We propose \model, a quantum-inspired framework for personalized federated fine-tuning with a compact shared adapter. Specifically, \model constructs reusable orthonormal adaptation bases from compact skew-symmetric generators using a truncated matrix-exponential approximation followed by QR orthonormalization. Compatible projections reuse these bases, while a globally shared full core preserves the flexibility of each projection. Each client further maintains a lightweight private residual operator, whose directions are learned globally and whose coefficients remain local. Experiments on seven public medical tasks and multiple LLM backbones show substantial reductions in trainable parameters and per-round adapter upload volume, with competitive predictive performance on most tasks.
\end{abstract}


\section{Introduction}
\label{sec:introduction}
\begin{figure}[t]
    \centering
    \includegraphics[width=\linewidth]{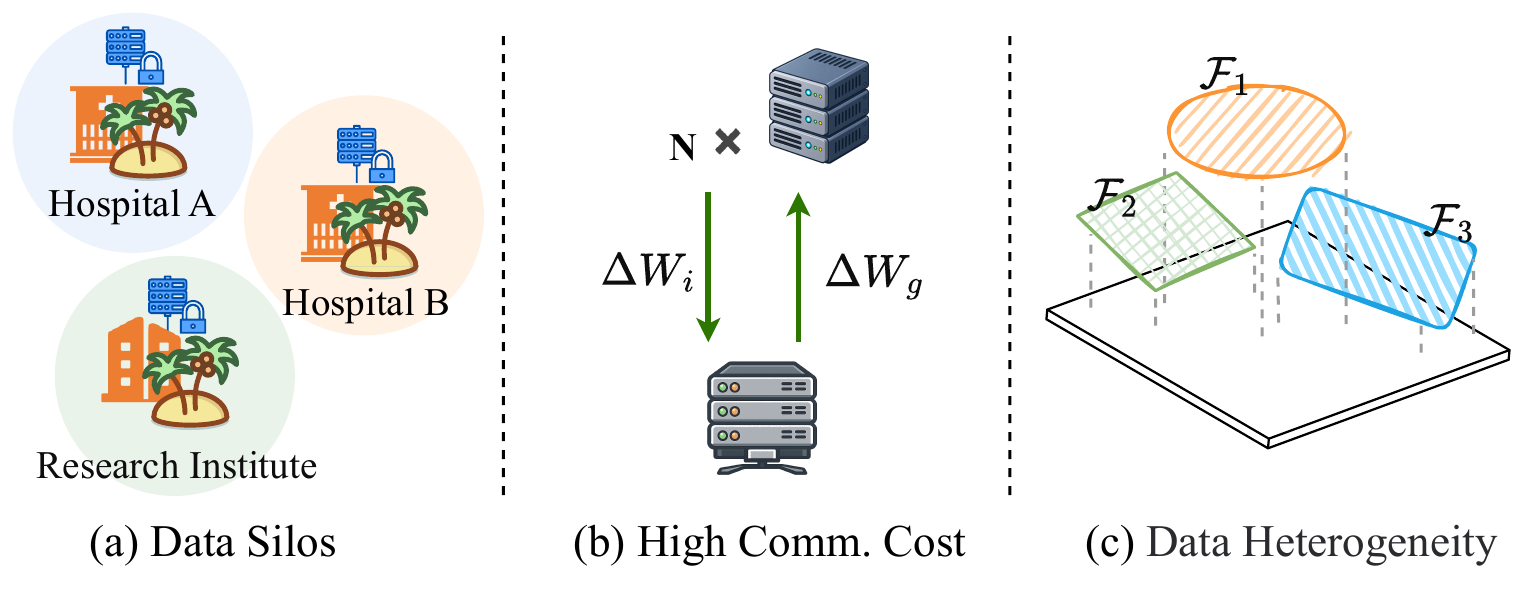}
    \caption{Challenges in FFT of medical LLMs: (a) institutional data silos, (b) high communication costs from repeated model update exchanges, and (c) heterogeneous data across institutions.}
    \label{fig:medical-fl-challenges}
\end{figure}

Medical institutions continuously generate large and heterogeneous volumes of clinical text, diagnostic records, annotated medical documents, and question answering data. These distributed resources provide valuable evidence for adapting large language models (LLMs) to a wide range of clinical and biomedical language processing tasks \cite{singhal2023clinical}. However, consolidating medical records into a centralized training repository is often impractical because of privacy regulations, data governance requirements, and institutional ownership constraints \cite{rieke2020digital}. Federated learning (FL) offers a practical alternative by enabling multiple institutions to collaborate while retaining sensitive records locally \cite{mcmahan2017fedavg,rieke2020digital}. We refer to adapting pretrained models under this distributed protocol as federated fine-tuning (FFT). Nevertheless, hospital servers and clinical edge devices often vary in computing power, memory capacity, and network bandwidth, making repeated optimization and transmission of LLMs prohibitively expensive \cite{li2020fedprox,zhang2023fedpetuning}. As illustrated in Fig.~\ref{fig:medical-fl-challenges}, institutional data silos motivate FFT, while high communication costs and heterogeneous local data call for efficient personalized federated adaptation.

In a typical FL workflow, the server distributes the current global parameters, clients optimize them on private local data, and the resulting updates are aggregated to improve the global model \cite{mcmahan2017fedavg}. Directly adapting all parameters of an LLM is often impractical because of the associated computation, memory, and communication costs. Parameter-efficient fine-tuning (PEFT) addresses this limitation by freezing the pretrained backbone and optimizing only a compact set of task parameters \cite{houlsby2019adapters,hu2021lora}. Within FFT, PEFT allows clients to train and exchange lightweight adaptation modules rather than the entire model, reducing local resource consumption and communication overhead while leaving the pretrained backbone weights unchanged \cite{zhang2023fedpetuning}. Low-Rank Adaptation (LoRA), for example, represents each weight update with two trainable low-rank factors \cite{hu2021lora}. However, these factors are typically learned independently for individual projection layers, and their trainable and communicated parameter counts still scale with the dimensions of the adapted modules. Moreover, the adapter payload is exchanged repeatedly over communication rounds, so even a relatively small fraction of the backbone parameters can accumulate into a substantial transmission volume. Reducing this recurring cost requires attention to how adaptation parameters are organized and reused across the model, in addition to how many parameters are trained in each individual module.

Despite these advantages, standard FFT with independently parameterized low-rank modules and fully shared aggregation presents two technical limitations. Conventional low-rank adapters learn a separate pair of projection factors for each adapted layer, requiring approximately \(r(d_{\mathrm{in}}+d_{\mathrm{out}})\) trainable parameters per weight matrix, where \(r\) is the adaptation rank and \(d_{\mathrm{in}}\) and \(d_{\mathrm{out}}\) are the input and output dimensions \cite{hu2021lora}. Independently parameterizing these factors across Transformer layers incurs repeated parameter costs, with limited reuse of adaptation directions across compatible projections. Simply lowering the adaptation rank reduces the parameter budget but also restricts the dimension of the update space. An alternative is to reuse adaptation bases across compatible projections while retaining transformations specific to each projection within those bases. This separates the cost of specifying adaptation directions from the cost of combining them for each projection.

Conventional FL aggregation generally treats all adapter parameters as globally shareable. When medical data are not independently and identically distributed (non-IID), client updates can contain both transferable clinical knowledge and institution specific information, leading to conflicting local objectives and unstable aggregation \cite{li2020fedprox,karimireddy2020scaffold}. Aggregating both components can weaken local specialization, whereas keeping complete adapters private restricts knowledge transfer across institutions \cite{arivazhagan2019fedper}. A lightweight private component offers a middle ground: clients can share the main adaptation parameters while retaining a small set of local corrections \cite{arivazhagan2019fedper,collins2021fedrep}. The challenge is to provide useful local flexibility without introducing a complete private adapter for every institution. Together, these limitations motivate two related challenges: constructing reusable low-rank adaptation subspaces from compact intrinsic parameters and separating globally collaborative transformations from lightweight client specific adaptation without substantially increasing the communication payload.

Quantum-inspired parameterizations offer a structured perspective on constructing adaptation transformations \cite{koike2025quantumpeft}. In quantum computation, unitary transformations preserve inner products and norms and can be expressed as matrix exponentials of Hermitian generators, \(\mathbf Q=\exp(-\mathrm{i}\mathbf H)\) \cite{wierichs2022general}. For real-valued adaptation, a corresponding construction uses a real skew-symmetric generator \(\mathbf A\), whose exponential \(\exp(\mathbf A)\) is orthogonal; setting \(\mathbf H=\mathrm{i}\mathbf A\) establishes the connection to the unitary formulation \cite{lezcano2019orthogonal}. Applying this orthogonal transformation to a set of standard basis vectors yields orthonormal directions that can serve as adaptation bases. The opportunity for parameter reduction comes from imposing a compact structure on the generator and reusing the resulting bases, rather than from orthogonality alone. This perspective motivates constructing shared adaptation subspaces through structured generator parameters while retaining separate core transformations for each projection and client-specific adjustments. 

To address these challenges, we propose Personalized Federated Quantum-Inspired Fine-Tuning (\model), an efficient method for adapting medical LLMs in distributed environments. Inspired by generator-based unitary transformations, the method uses structured skew-symmetric generators and matrix exponential mappings to construct orthogonal adaptation subspaces \cite{koike2025quantumpeft}. A truncated Taylor approximation followed by thin QR orthonormalization produces the adaptation bases without explicitly forming a dense matrix exponential. Projections with matching module types and matrix shapes reuse these bases, while a full core matrix for each projection retains flexibility in combining the shared directions. All computations and optimization are performed on classical hardware.

For institutional personalization, \model combines a globally aggregated full core with a private residual operator for each client. Globally learned residual direction matrices map the client retained coefficient vector to a structured core correction, allowing each institution to adjust selected directions within the shared adaptation subspaces. This design introduces only \(r\) private parameters per adapted projection, compared with \(r^2\) for a full private core. Clients retain these coefficients across communication rounds and exchange only the shared adapter parameters. By coupling basis reuse with compact local corrections, the framework is designed to reduce the trainable parameter budget while supporting both collaborative learning and institution specific adaptation. The main contributions are summarized as follows:
\begin{itemize}
    \item A quantum-inspired FFT framework is formulated to adapt medical LLMs across clients with heterogeneous data and constrained resources.
    \item Orthonormal adaptation bases are constructed from compact skew-symmetric generators using a truncated matrix-exponential approximation followed by thin QR orthonormalization, and are shared across compatible Transformer projections.
    \item A shared and private core decomposition combines an aggregated full core with structured residual operators controlled by locally retained coefficients, preserving institutional adaptation without additional communication payload.
    \item Experiments on medical datasets demonstrate that \model substantially reduces trainable parameters while maintaining competitive performance.
\end{itemize}

\section{Related Work}
\label{sec:related-work}

\subsection{Parameter-Efficient Fine-Tuning}
\label{sec:peft}
PEFT adapts a pretrained model by freezing most backbone weights and optimizing a small set of task parameters. Bottleneck adapters insert compact trainable modules between frozen Transformer components \cite{houlsby2019adapters}, whereas LoRA represents each weight update with two trainable low-rank factors \cite{hu2021lora}. Recent studies have refined LoRA from several complementary perspectives. LoRA+ assigns different learning rates to its two factors to improve optimization efficiency \cite{hayou2024loraplus}, while DoRA decomposes pretrained weights into magnitude and direction components and applies low-rank updates to the directional component \cite{liu2024dora}. RandLoRA constructs full-rank updates from learned combinations of fixed random low-rank bases \cite{albert2025randlora}, and LowRA combines LoRA with quantization below two bits per parameter to reduce memory consumption \cite{zhou2025lowra}. PLoP further studies where adapters should be placed and automatically selects projection types for a given model and task \cite{hayou2026plop}. Efficient foundation model adaptation has also been examined in both centralized and federated settings \cite{saadati2025foundation}. Despite this progress, most designs retain separate parameters for individual layers or projections. For a projection of shape $d_o\times d_i$, LoRA requires $r(d_o+d_i)$ parameters, so repeatedly adapting multiple Transformer projections limits structural reuse and preserves dimension dependent training and communication costs.

\subsection{Federated Learning}
\label{sec:fl}
FL enables distributed clients to collaboratively optimize models while retaining training data locally. FedAvg alternates local optimization with aggregation weighted by client sample counts \cite{mcmahan2017fedavg}. Recent work addresses statistical heterogeneity through adaptive aggregation and personalized collaboration. FedAWA adjusts aggregation weights using client update directions to improve alignment with global optimization \cite{shi2025fedawa}. DPFL constructs a collaboration graph through constrained greedy selection, enabling decentralized personalized training within predefined communication and resource budgets \cite{kharrat2025dpfl}. FedMerge jointly learns multiple global models and merging weights for each client to produce personalized models suited to local data distributions \cite{chen2026fedmerge}. These approaches improve how clients collaborate, but do not directly address the repeated parameterization of adaptation modules across Transformer projections. For medical LLM adaptation, a complementary challenge is to couple compact reusable subspaces with lightweight institutional corrections, rather than relying solely on changes to aggregation or collaboration structure.

\subsection{Federated Fine-Tuning}
\label{sec:fed-peft}
Parameter-efficient FFT integrates PEFT into FL, allowing clients to retain raw data and communicate only a small trainable parameter subset. FedPer introduced shared and private layers for personalized FL \cite{arivazhagan2019fedper}. FedPETuning evaluated multiple PEFT methods under federated constraints \cite{zhang2023fedpetuning}. FedPepTAO combined partial prompt tuning with adaptive optimization to mitigate client drift \cite{che2023fedpeptao}. FFA-LoRA freezes the randomly initialized nonzero factor and updates only the zero initialized factor \cite{sun2024ffalora}. FedEx-LoRA introduces a residual correction to recover the exact aggregated update \cite{singhal2025fedex}. LoRA-FAIR jointly refines server aggregation and client initialization \cite{bian2025lorafair}. RoLoRA alternates optimization of the two LoRA factors \cite{chen2025rolora}. RAFFT uses Riemannian parameter matching to support adaptive ranks \cite{zhou2026rafft}. FedALT combines individual adapters with a rest-of-world adapter for personalized training \cite{bian2026fedalt}. Personalized FL has also been investigated for heterogeneous medical imaging systems \cite{shadin2025heterogeneity}. Nevertheless, these approaches primarily correct or personalize complete LoRA modules. They do not directly provide reusable orthogonal subspaces across compatible projections together with a shared and private core decomposition for heterogeneous medical language modeling.


\section{Methodology}
\label{sec:methodology}

\begin{figure*}[!ht]
    \centering
    \includegraphics[width=1.0\textwidth]{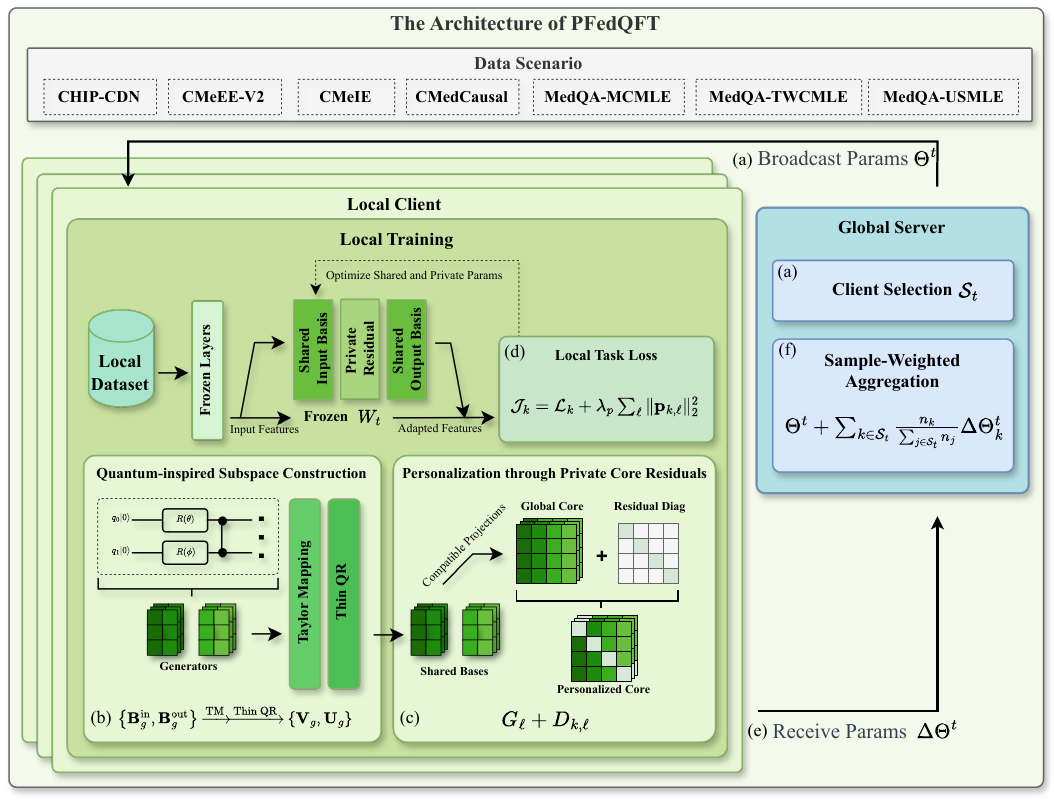}
    \caption{Overall architecture of PFedQFT. The framework comprises a global server that coordinates sample-weighted aggregation and multiple clients that construct reusable adaptation subspaces while retaining private core residuals for personalized local training. \textbf{(a)} At communication round $t$, the server selects the participating clients $\mathcal S_t$ and broadcasts the shared parameters $\Theta^t$. \textbf{(b)} Each selected client transforms the generators $\mathbf B_g^{\mathrm{in}}$ and $\mathbf B_g^{\mathrm{out}}$ associated with sharing group $g$ into the orthonormal bases $\mathbf V_g$ and $\mathbf U_g$ through Taylor mapping and thin QR. \textbf{(c)} For each adapted projection $\ell$, client $k$ constructs the private residual operator $\mathbf D_{k,\ell}=\mathbf R_\ell^{\mathrm{out}}\operatorname{diag}(\mathbf p_{k,\ell})(\mathbf R_\ell^{\mathrm{in}})^\top$ and the personalized subspace operator $\mathbf C_{k,\ell}=\mathbf G_\ell+\mathbf D_{k,\ell}$. \textbf{(d)} Local optimization updates the shared parameter copy $\Theta_k$ and the private coefficients $\phi_k$ using $\mathcal D_k$. \textbf{(e)} Client $k$ uploads only $(\Delta\Theta_k^t,n_k)$, while $\mathcal D_k$ and $\mathbf p_{k,\ell}$ remain local. \textbf{(f)} The server performs sample-weighted aggregation to obtain the shared parameters $\Theta^{t+1}$ for the next communication round.}
    \label{fig:pfedqft-framework}
\end{figure*}

Fig.~\ref{fig:pfedqft-framework} outlines the interaction between server-side coordination and client-side adaptation in \model. On each client, quantum-inspired subspace construction generates reusable orthonormal bases, while private core residuals introduce institution-specific adaptation within the shared subspace. The server manages client participation and aggregates only shared parameter updates, thereby separating collaborative optimization from locally retained data and personalization parameters. The following subsections formalize these components.

\subsection{Problem Formulation}
\label{sec:problem-formulation}
Consider a federated system comprising $K$ medical institutions, each represented by a client. Client $k$ maintains a private dataset $\mathcal D_k$ containing $n_k$ samples, with $N=\sum_{k=1}^{K}n_k$ denoting the total number of samples. Throughout this paper, personalization is defined at the institutional client level: the private parameters $\phi_k$ adapt the model to institution $k$ and its local data distribution, rather than to an individual patient. Given a pretrained language model with frozen parameters $W$, adaptation is performed through globally shared parameters $\Theta$ and private parameters $\phi_k$ specific to client $k$. The personalized federated learning objective is formulated as:
\begin{equation}
 \min_{\Theta,\{\phi_k\}_{k=1}^{K}}
 \sum_{k=1}^{K}\frac{n_k}{N}
 \mathcal L_k(W,\Theta,\phi_k),
 \label{eq:federated-objective}
\end{equation}
where $\mathcal L_k$ denotes the empirical task loss evaluated on $\mathcal D_k$. Under this protocol, raw medical records and private adaptation parameters remain at their respective clients, while only globally shared adaptation parameters are exchanged with the server. For each selected projection matrix $\mathbf{W}_\ell\in\mathbb R^{d_o\times d_i}$, where $d_i$ and $d_o$ denote its input and output dimensions, respectively, client $k$ uses the adapted matrix $\mathbf{W}_\ell+\Delta \mathbf{W}_{k,\ell}(\Theta,\phi_k)$. The objective is to construct a compact parameterization of $\Delta \mathbf{W}_{k,\ell}$ that reduces trainable parameter counts and communication overhead relative to full fine-tuning while supporting knowledge sharing across institutions and institution specific adaptation.

\subsection{Quantum-Inspired Subspace Adaptation}
\label{sec:quantum-subspace}
To enable parameter sharing across compatible Transformer projections, compact trainable parameters are used to construct orthonormal bases for their adaptation subspaces. The pretrained backbone remains frozen, with lightweight adapters introduced into selected attention and feed forward projections. Let $\mathcal L$ denote the set of adapted projections and $\mathcal G$ the set of sharing groups. Projections with the same module type and matrix shape form a group $g\in\mathcal G$, and $g(\ell)$ maps projection $\ell\in\mathcal L$ to its group. Each group has input and output dimensions $d_{i,g}$ and $d_{o,g}$, respectively, and uses bases $\mathbf U_g\in\mathbb R^{d_{o,g}\times r}$ and $\mathbf V_g\in\mathbb R^{d_{i,g}\times r}$, with $r\leq\min(d_{i,g},d_{o,g})$. Both bases have orthonormal columns and are generated from parameters shared within the group.

Inspired by parameterized quantum gates that control state evolution through unitary transformations, the method generates adaptation bases from compact transformation parameters \cite{koike2025quantumpeft}. Real orthogonal mappings implement this design, producing adaptation directions shared across compatible projections. For a feature dimension $d$, let $\kappa$ control the capacity of the structured generator, with $1\leq\kappa\leq d$. Define $\mathbf{P}_\kappa=[\mathbf{e}_1,\ldots,\mathbf{e}_\kappa]\in\mathbb R^{d\times\kappa}$ using the first $\kappa$ standard basis vectors. A trainable masked matrix $\mathbf{B}\in\mathbb R^{d\times\kappa}$ satisfies $[\mathbf{B}]_{ij}=0$ for $i\leq j$. The resulting generator and its skew-symmetry property are given by:
\begin{align}
 \mathbf{A}(\mathbf{B})&=\mathbf{B}\mathbf{P}_\kappa^\top-\mathbf{P}_\kappa \mathbf{B}^\top,
 \label{eq:skew-generator}\\
 \mathbf{A}(\mathbf{B})^\top&=-\mathbf{A}(\mathbf{B}).
 \label{eq:skew-symmetry}
\end{align}

Its exponential $\mathbf{Q}(\mathbf{B})=\exp(\mathbf{A}(\mathbf{B}))$ is orthogonal and represents a real valued instance of a unitary transformation. In particular, $\mathbf{H}(\mathbf{B})=\mathrm{i}\mathbf{A}(\mathbf{B})$ is Hermitian and $\mathbf{Q}(\mathbf{B})=\exp(-\mathrm{i}\mathbf{H}(\mathbf{B}))$, linking the construction to the generator formulation of quantum state evolution. Applying $\mathbf{Q}(\mathbf{B})$ to $\mathbf{E}_r=[\mathbf{e}_1,\ldots,\mathbf{e}_r]\in\mathbb R^{d\times r}$ selects $r$ orthonormal adaptation directions. To compute these directions efficiently, a degree-$J$ Taylor approximation is applied directly to $\mathbf{E}_r$, followed by thin QR to obtain the orthonormal adaptation basis:
\begin{align}
 \widetilde{\mathbf{F}}_d(\mathbf{B})&=\sum_{j=0}^{J}\frac{\mathbf{A}(\mathbf{B})^j\mathbf{E}_r}{j!},
 \label{eq:taylor-approximation}\\
 \mathbf{F}_d(\mathbf{B})&=\mathcal Q\!\left(\widetilde{\mathbf{F}}_d(\mathbf{B})\right),
\label{eq:taylor-frame}
\end{align}
where $\mathcal Q(\cdot)$ denotes the operator that returns the orthonormal factor of a thin QR decomposition, distinct from the matrix exponential $\mathbf{Q}(\mathbf{B})$. The order $J$ controls the truncation depth. Provided that $\widetilde{\mathbf{F}}_d(\mathbf{B})$ has full column rank, thin QR yields an orthonormal basis for its column space. Each multiplication is evaluated as $\mathbf{A}(\mathbf{B})\mathbf{Z}=\mathbf{B}(\mathbf{P}_\kappa^\top \mathbf{Z})-\mathbf{P}_\kappa(\mathbf{B}^\top \mathbf{Z})$ for an intermediate matrix $\mathbf{Z}$, avoiding construction of a dense $d\times d$ generator or exponential. For group $g$, separate masked generators $\mathbf B_g^{\mathrm{in}}\in\mathbb R^{d_{i,g}\times\kappa}$ and $\mathbf B_g^{\mathrm{out}}\in\mathbb R^{d_{o,g}\times\kappa}$ construct the input and output bases:
\begin{align}
 \mathbf V_g&=\mathbf F_{d_{i,g}}(\mathbf B_g^{\mathrm{in}}),
 \label{eq:input-basis}\\
 \mathbf U_g&=\mathbf F_{d_{o,g}}(\mathbf B_g^{\mathrm{out}}).
 \label{eq:output-basis}
\end{align}

These bases satisfy $\mathbf V_g^\top\mathbf V_g=\mathbf I_r$ and $\mathbf U_g^\top\mathbf U_g=\mathbf I_r$. They are computed once per model call and reused within each sharing group, avoiding a separately parameterized basis for every layer.

\subsection{Personalized Federated Adaptation with Private Core Residuals}
\label{sec:personalized-spectral}
Personalized federated adaptation requires local flexibility within the shared adaptation subspaces. Each adapted projection $\ell$ therefore uses a globally shared full core $\mathbf G_\ell\in\mathbb R^{r\times r}$, which captures interactions between all input and output adaptation directions. Unlike the orthonormal bases reused across compatible projections, $\mathbf G_\ell$ is specific to projection $\ell$ and is shared across clients through federated aggregation.

For the private component, two globally shared residual direction matrices $\mathbf R_\ell^{\mathrm{in}},\mathbf R_\ell^{\mathrm{out}}\in\mathbb R^{r\times r}$ satisfy $(\mathbf R_\ell^{s})^\top\mathbf R_\ell^{s}=\mathbf I_r$ for $s\in\{\mathrm{in},\mathrm{out}\}$. Each matrix is block diagonal with $r/h$ blocks of size $h\times h$, where $2\leq h\leq r$ and $h$ divides $r$. Within each block, an identity matrix is successively right multiplied by Givens rotations for all coordinate pairs $(a,b)$ with $1\leq a<b\leq h$, in lexicographic order. Each rotation uses the $2\times2$ submatrix $[\cos\rho,-\sin\rho;\sin\rho,\cos\rho]$ on its selected coordinates. The resulting angle vector $\boldsymbol\rho_\ell^{s}\in\mathbb R^{r(h-1)/2}$ is initialized to zero, yielding an identity direction matrix. This construction learns residual directions within each block without mixing coordinates across blocks.

The direction angles are aggregated together with the full core and subspace generator parameters, whereas client $k$ retains only a coefficient vector $\mathbf p_{k,\ell}\in\mathbb R^r$. The private residual operator, personalized subspace operator, and resulting update to the frozen projection are defined as:
\begin{align}
 \mathbf D_{k,\ell}
 &=\mathbf R^{\mathrm{out}}_\ell
 \operatorname{diag}(\mathbf p_{k,\ell})
 (\mathbf R^{\mathrm{in}}_\ell)^\top,
 \label{eq:private-residual-operator}\\
 \mathbf C_{k,\ell}
 &=\mathbf G_\ell+\mathbf D_{k,\ell},
 \label{eq:personalized-core}\\
 \Delta\mathbf W_{k,\ell}
 &=\frac{\alpha}{r}
 \mathbf U_{g(\ell)}\mathbf C_{k,\ell}\mathbf V_{g(\ell)}^\top,
 \label{eq:fed-tqft-update}
\end{align}
where $g(\ell)$ identifies the sharing group of projection $\ell$ and $\alpha$ controls the update scale.

The full core $\mathbf G_\ell$ provides a complete shared transformation within the adaptation subspaces, whereas $\mathbf D_{k,\ell}$ supplies a structured institution specific correction. For fixed residual direction matrices, the client retained coefficients control $r$ learned rank one directions rather than only the coordinate axes of the core. The private component consequently uses only $r$ client parameters per projection, compared with $r^2$ for a full private core. These coefficients remain at their respective clients and do not increase the shared adapter payload. Global cores are initialized with small nonzero values to permit task gradients to reach the basis generators at the start of training. Private coefficients are initialized to zero, so the initial private residual vanishes. Under this parameterization, the task gradient with respect to the residual direction angles is zero whenever $\mathbf p_{k,\ell}=\mathbf 0$. The private coefficients can nevertheless receive nonzero task gradients; once they become nonzero, the direction angles can also receive task gradients. A nonzero global core does not directly provide task gradients to these angles, since they appear only in the private branch.

\subsection{Federated Optimization}
\label{sec:federated-optimization}
The globally shared and private parameters are organized as tuples:
\begin{align}
 \Theta&=\Bigl(
   (\mathbf B_g^{\mathrm{in}},\mathbf B_g^{\mathrm{out}})_{g\in\mathcal G},
   \notag\\[-2pt]
 &\qquad
   (\mathbf G_\ell,\boldsymbol\rho_\ell^{\mathrm{in}},
   \boldsymbol\rho_\ell^{\mathrm{out}})_{\ell\in\mathcal L}
   \Bigr),
 \label{eq:shared-parameter-tuple}\\
 \phi_k&=(\mathbf p_{k,\ell})_{\ell\in\mathcal L}.
 \label{eq:private-parameter-tuple}
\end{align}

Differences and weighted averages of $\Theta$ act componentwise on the corresponding tensors. At round $t$, the server selects a client subset $\mathcal S_t$ and broadcasts $\Theta^t$. Each participating client initializes its shared parameters from $\Theta^t$ and restores its previously retained private coefficients $\phi_k^t$. In practice, training uses a regularized version of the objective in Eq.~\eqref{eq:federated-objective}, replacing each $\mathcal L_k$ with $\mathcal J_k$. The client performs $E$ local epochs of minibatch optimization using the local objective:
\begin{equation}
 \mathcal J_k(\Theta,\phi_k)
 =\mathcal L_k(W,\Theta,\phi_k)
 +\lambda_p\sum_{\ell\in\mathcal L}\|\mathbf p_{k,\ell}\|_2^2,
 \label{eq:local-objective}
\end{equation}
where $\lambda_p\geq0$ controls the regularization of private residual coefficients. Gradients update both shared and private parameters through the basis construction and core transformations, yielding $(\Theta_k^{t+1},\phi_k^{t+1})$ after the prescribed local epochs.

Client $k$ retains $\phi_k^{t+1}$ locally and uploads its shared parameter increment with the sample count $n_k$. The increment and the server update are defined as:
\begin{align}
 \Delta\Theta_k^t&=\Theta_k^{t+1}-\Theta^t,\label{eq:adapter-increment}\\
 \Theta^{t+1}&=\Theta^t+\sum_{k\in\mathcal S_t}
 \frac{n_k}{\sum_{j\in\mathcal S_t}n_j}\Delta\Theta_k^t.
 \label{eq:fedavg-adapter}
\end{align}

Aggregation operates on the trainable generator parameters and residual direction angles, together with the globally shared full cores. The orthonormal bases and residual direction matrices are reconstructed from the updated parameters using their respective mappings. Private coefficients do not enter the server update, and clients outside $\mathcal S_t$ retain their previous state:
\begin{equation}
 \phi_k^{t+1}=\phi_k^t,\qquad k\notin\mathcal S_t.
 \label{eq:private-state-retention}
\end{equation}

Algorithm~\ref{alg:fed-tqft} summarizes the basis construction, local training, and federated aggregation procedure.

\begin{algorithm}[!h]
\caption{Federated Optimization}
\label{alg:fed-tqft}
\small
\SetInd{0.4em}{0.6em}
\Input{Frozen LLM $W$; datasets $\{\mathcal D_k\}_{k=1}^{K}$; sharing groups $\mathcal G$; ranks $r,\kappa$; Taylor order $J$; rounds $T$; local epochs $E$; update scale $\alpha$; regularization coefficient $\lambda_p$}
\Output{Global adapter $\Theta^T$ and locally retained $\{\phi_k^T\}_{k=1}^{K}$}
\BlankLine
Initialize shared generators, residual direction angles, and full cores as $\Theta^0$\;
Initialize private coefficients $\phi_k^0=\{\mathbf p_{k,\ell}^0=\mathbf 0\}_\ell$\;
\For{$t=0$ \KwTo $T-1$}{
    Select $\mathcal S_t$ and broadcast $\Theta^t$\;
    Keep $\phi_k^{t+1}\gets\phi_k^t$ for $k\notin\mathcal S_t$\;
    \ForEach{$k\in\mathcal S_t$ \textbf{in parallel}}{
        Load $\Theta_k\gets\Theta^t$ and restore $\phi_k\gets\phi_k^t$\;
        \ForEach{minibatch $\mathcal B\subset\mathcal D_k$ over $E$ local epochs}{
            \algstage{Shared Subspace Construction}
            \ForEach{input and output generator $\mathbf B$ in $\mathcal G$}{
                Apply the lower triangular mask to $\mathbf B$\;
                $\mathbf Z\gets\mathbf E_r$; $\widetilde{\mathbf F}\gets\mathbf E_r$\;
                \For{$j=1$ \KwTo $J$}{
                    $\mathbf Z\gets[\mathbf B(\mathbf P_\kappa^\top\mathbf Z)-\mathbf P_\kappa(\mathbf B^\top\mathbf Z)]/j$\;
                    $\widetilde{\mathbf F}\gets\widetilde{\mathbf F}+\mathbf Z$\;
                }
                Store $\mathcal Q(\widetilde{\mathbf F})$ as $\mathbf V_g$ or $\mathbf U_g$\;
            }
            \algstage{Core Adaptation}
            \ForEach{adapted projection $\ell$}{
                Construct $\mathbf R_\ell^{\mathrm{in}},\mathbf R_\ell^{\mathrm{out}}$ from residual direction angles\;
                $\mathbf D_{k,\ell}\gets\mathbf R_\ell^{\mathrm{out}}\operatorname{diag}(\mathbf p_{k,\ell})(\mathbf R_\ell^{\mathrm{in}})^\top$\;
                $\mathbf C_{k,\ell}\gets\mathbf G_\ell+\mathbf D_{k,\ell}$\;
                Apply $\Delta\mathbf W_{k,\ell}$ using Eq.~\eqref{eq:fed-tqft-update}\;
            }
            Evaluate minibatch loss $\mathcal J_k$ using Eq.~\eqref{eq:local-objective}\;
            Update $\Theta_k,\phi_k$ by backpropagation through $\mathcal J_k$\;
        }
        Store $\phi_k^{t+1}\gets\phi_k$ locally\;
        Upload $\Delta\Theta_k^t\gets\Theta_k-\Theta^t$ and $n_k$\;
    }
    \algstage{Global Aggregation}
    $\displaystyle\Theta^{t+1}\gets\Theta^t+
    \sum_{k\in\mathcal S_t}\frac{n_k}{\sum_{j\in\mathcal S_t}n_j}\Delta\Theta_k^t$\;
}
\Return $\Theta^T$; each client retains its own $\phi_k^T$\;
\end{algorithm}

\section{Experiments}
\label{sec:experiments}

\subsection{Datasets and Models}
\label{sec:datasets}
Experiments are conducted on seven medical datasets covering diagnosis normalization, information extraction, and medical question answering. The Chinese biomedical tasks are drawn from CBLUE~\cite{zhang2022cblue} and its subsequent releases: CHIP-CDN for diagnosis normalization, CMeEE-V2 for named entity recognition, CMeIE for entity relation extraction, and CMedCausal~\cite{li2022cmedcausal} for causal relation extraction. Medical question answering is evaluated on the MedQA-MCMLE, MedQA-TWCMLE, and MedQA-USMLE subsets~\cite{jin2021medqa}, which contain medical licensing examination questions in simplified Chinese, traditional Chinese, and English, respectively. Each MedQA subset is evaluated as a separate task.

We evaluate \model using five backbone LLMs: Llama-3.2-3B-Instruct, Llama-3.1-8B-Instruct~\cite{grattafiori2024llama3}, Qwen3-1.7B, Qwen3-4B~\cite{yang2025qwen3}, and MedGemma-1.5-4B-it~\cite{sellergren2026medgemma15}. These models are selected to assess the applicability of \model across model families and parameter scales, encompassing both general-purpose and medically adapted backbones.

\subsection{Baselines and Metrics}
\label{sec:baselines-models}
We compare \model with four FFT baselines based on LoRA, covering exact aggregation through factor stacking, adaptive server optimization, factor freezing, and residual correction. FLoRA performs noise-free aggregation by vertically stacking the sample-weighted local $\mathbf A_k$ factors and horizontally stacking the corresponding $\mathbf B_k$ factors, so that the product of the global factors exactly equals the sample-weighted sum of the local LoRA updates; the stacking construction also supports heterogeneous local ranks~\cite{wang2024flora}. FedAdam-LoRA applies the FedAdam server optimizer to the aggregated LoRA parameter updates~\cite{reddi2021adaptive}. FFA-LoRA fixes the randomly initialized factor $\mathbf A$ and trains only $\mathbf B$, reducing communication and eliminating the aggregation mismatch induced by independently averaging two trainable factors~\cite{sun2024ffalora}. FedEx-LoRA introduces a residual correction to recover the weighted average of client weight updates from the aggregated low-rank factors~\cite{singhal2025fedex}.

\label{sec:experimental-metrics}
Predictive performance is assessed using task-specific metrics. For medical diagnosis normalization and information extraction datasets (CHIP-CDN, CMeEE-V2, CMeIE, and CMedCausal), we adopt Soft F1, a similarity-weighted micro F1 score that awards partial credit for lexical overlap. Predictions and references are parsed into sets of diagnosis terms or structured extraction units. For each example $i$, the Hungarian algorithm~\cite{kuhn1955hungarian} identifies a maximum-weight one-to-one matching $\mathcal M_i$ between the predicted set $\widehat{\mathcal Y}_i$ and reference set $\mathcal Y_i$. Given unit similarity $s(u,v)\in[0,1]$, the score is computed over the evaluation set as:
\begin{equation}
\operatorname{SoftF1}=
\frac{2\sum_i\sum_{(u,v)\in\mathcal M_i}s(u,v)}
{\sum_i|\widehat{\mathcal Y}_i|+\sum_i|\mathcal Y_i|}.
\label{eq:soft-f1}
\end{equation}

For the three MedQA subsets, we report accuracy based on exact matching of predicted and reference answer option identifiers.

Parameter efficiency is measured by the number of trainable scalar parameters per client:
\begin{equation}
\nu_k=|\Theta|+|\phi_k|,
\label{eq:parametercount}
\end{equation}
where $|\cdot|$ counts scalar parameters, with shared tensors counted only once. The upload volume for each client in round $t$, measured in bytes, is:
\begin{equation}
\Gamma_{k,t}^{\uparrow}=\frac{b}{8}m_{k,t}^{\uparrow},
\label{eq:communication-volume}
\end{equation}
where $b$ is the number of bits per transmitted element and $m_{k,t}^{\uparrow}$ counts uploaded elements. Private residuals, protocol overhead, and initial backbone distribution are excluded.

\subsection{Configurations}
\label{sec:configurations}
All experiments freeze the pretrained backbone and optimize only the adaptation parameters, with Llama-3.2-3B-Instruct used by default unless otherwise specified. The LoRA-based baselines use a local rank of 16 at each client. FLoRA is evaluated with identical local ranks: every participating client trains a rank-16 adapter, and the server stacks the local factors so that the rank of the resulting global adapter equals the sum of the participating clients' local ranks. In contrast, \model aggregates shared adapter parameters across participating clients and retains institution specific residual coefficients locally. Each dataset is converted into a unified instruction--response format: diagnosis normalization produces standardized terms, information extraction tasks use structured textual outputs, and MedQA predicts an answer option identifier. Federated clients are simulated by partitioning the training records of each dataset. All experiments are conducted on a server equipped with two NVIDIA H100 GPUs, each with 80~GB of memory; the quantum-inspired operations in \model are implemented entirely on classical hardware.
\subsection{Experimental Results}

\begin{figure*}[!ht]
\centering
\includegraphics[width=\textwidth]{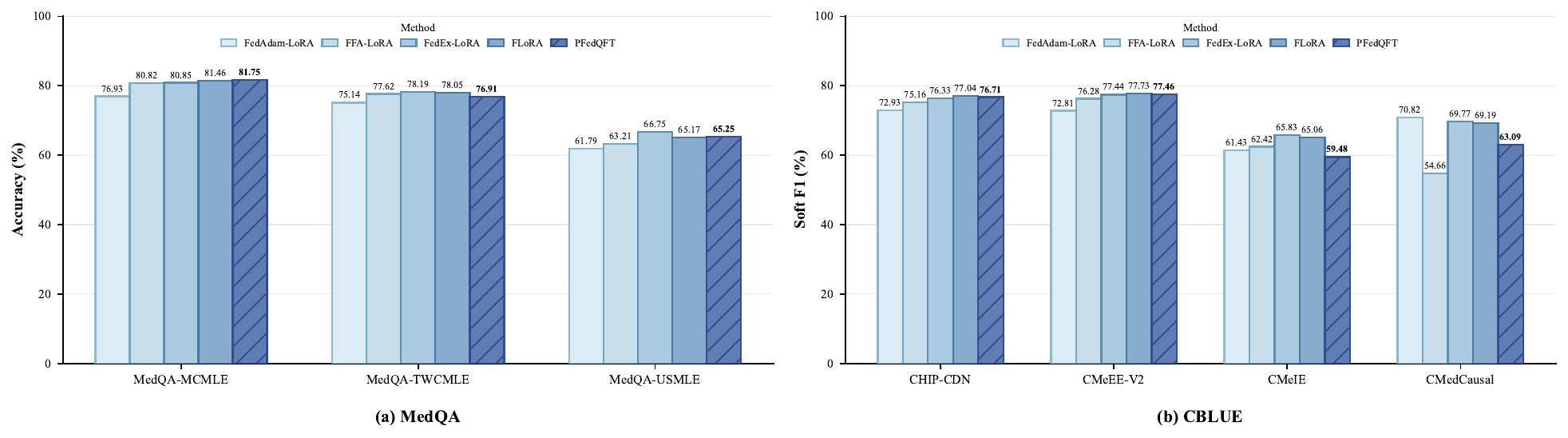}
\caption{Overall performance comparison. The Llama-3.2-3B-Instruct backbone is frozen, and only adapter parameters are optimized with five participating clients per communication round. The LoRA baselines use local rank-16 adapters, whereas \model uses task-specific adaptation configurations with globally shared parameters and residual coefficients retained privately by each client. Results are evaluated on the complete validation sets using Accuracy for medical question answering and Soft F1 for diagnosis normalization and information extraction.}
\label{fig:overall-performance}
\end{figure*}

\begin{figure*}[!ht]
\centering
\includegraphics[width=\textwidth]{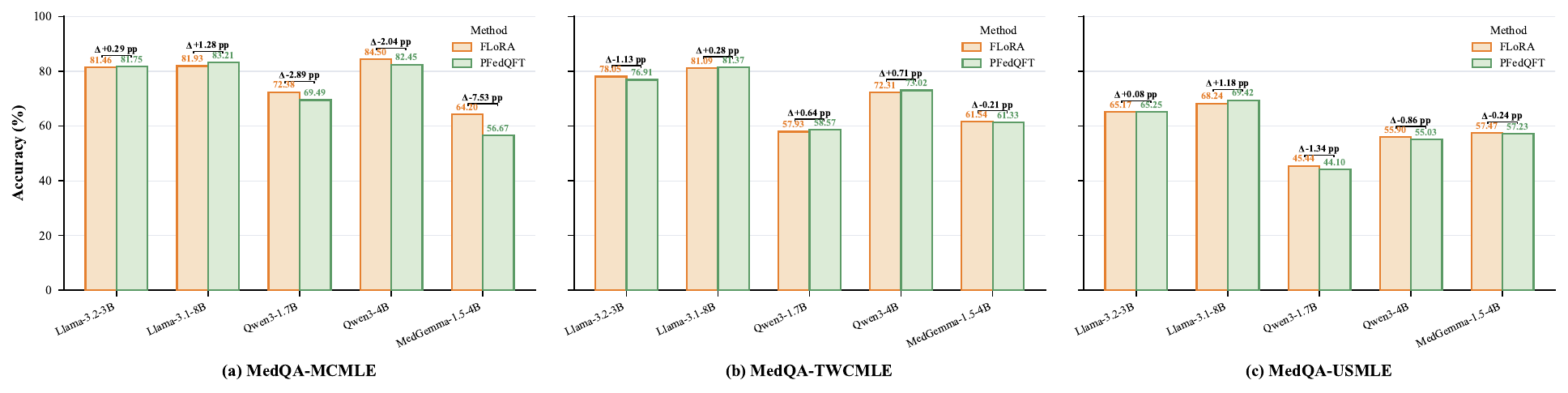}
\caption{Performance across backbone models. All backbones are frozen, and only adapter parameters are optimized with one local epoch per communication round, an effective batch size of 16, and a maximum sequence length of 768. FLoRA uses rank-16 adapters, whereas \model uses an adaptation rank of 8 and an intrinsic rank of 4 with residual coefficients retained privately by each client. Accuracy is evaluated on the complete validation splits, and $\Delta$ denotes the accuracy of \model minus that of FLoRA in percentage points.}
\label{fig:backbone-performance}
\end{figure*}

\begin{table}[!ht]
\centering
\caption{Performance under varying degrees of data heterogeneity.}
\label{tab:heterogeneity}
\setlength{\tabcolsep}{3pt}
\footnotesize
\begin{tabular*}{0.98\linewidth}{@{\extracolsep{\fill}}lrrrrrr@{}}
\toprule
\multirow{2}{*}{\shortstack{Non-IID\\Degree}}
& \multicolumn{2}{c}{MedQA-MCMLE}
& \multicolumn{2}{c}{CHIP-CDN}
& \multicolumn{2}{c}{CMeEE-V2} \\
\cmidrule(lr){2-3}
\cmidrule(lr){4-5}
\cmidrule(lr){6-7}
& FLoRA & \model
& FLoRA & \model
& FLoRA & \model \\
\midrule
Light & 81.83 & 81.49 & 78.32 & 76.52 & 77.79 & 76.74 \\
Moderate & 81.72 & 81.46 & 79.95 & 76.57 & 77.12 & 76.35 \\
Severe & 81.15 & 81.14 & 77.96 & 76.08 & 77.19 & 76.56 \\
\bottomrule
\end{tabular*}
\par\vspace{4pt}\noindent
\parbox{0.98\linewidth}{\scriptsize
\textit{Note:} The Dirichlet concentration parameter $\beta$ controls the non-IID degree, with $\beta=1.0$, $0.5$, and $0.1$ denoting light, moderate, and severe heterogeneity, respectively. MedQA-MCMLE uses Accuracy, while CHIP-CDN and CMeEE-V2 use Soft F1.
}
\end{table}

\begin{table*}[!ht]
\centering
\caption{Average trainable parameter count, adapter storage, communication cost, and runtime across the seven tasks.}
\label{tab:efficiency}
\small
\setlength{\tabcolsep}{3.5pt}
\begin{tabular*}{\textwidth}{@{\extracolsep{\fill}}lrrrrrr@{}}
\toprule
Method & \#TP (M) & TP (\%) & Shared (MB) & Private (MB) & Upload (MB) & Time \\
\midrule
FLoRA$^{\dagger}$ & 24.3139 & 0.7568 & 97.2554 & -- & 97.2554 & 0 h 45 min \\
FedAdam-LoRA & 24.3139 & 0.7568 & 97.2554 & -- & 97.2554 & 0 h 37 min \\
FFA-LoRA & 12.3863 & 0.3855 & 49.5452 & -- & 49.5452 & 1 h 06 min \\
FedEx-LoRA$^{\ddagger}$ & 24.3139 & 0.7568 & 97.2554 & -- & 97.2554 & 0 h 52 min \\
\model & 1.6524 & 0.0514 & 6.5981 & 0.011648 & 6.5981 & 6 h 23 min \\
\bottomrule
\end{tabular*}
\par\vspace{4pt}\noindent
\parbox{\textwidth}{\scriptsize
\textit{Note:} Values are arithmetic means across the seven task-specific experimental runs underlying the overall comparison in Fig.~\ref{fig:overall-performance}. Time denotes end-to-end wall-clock runtime, including training and evaluation. M denotes $10^6$ parameters, and MB denotes $10^6$ bytes. All adapter tensors are stored in FP32. \#TP denotes the trainable parameter count, and TP (\%) is normalized by the backbone parameter count. The Shared, Private, and Upload columns denote the shared adapter size, private storage for each client, and upload volume for each client per round, respectively.
$\dagger$ FLoRA uses aggregation based on factor stacking and is evaluated with a local rank of 16 at every participating client.
$\ddagger$ FedEx-LoRA excludes residual-correction traffic.
}
\end{table*}

\begin{figure*}[!ht]
\centering
\includegraphics[width=\textwidth]{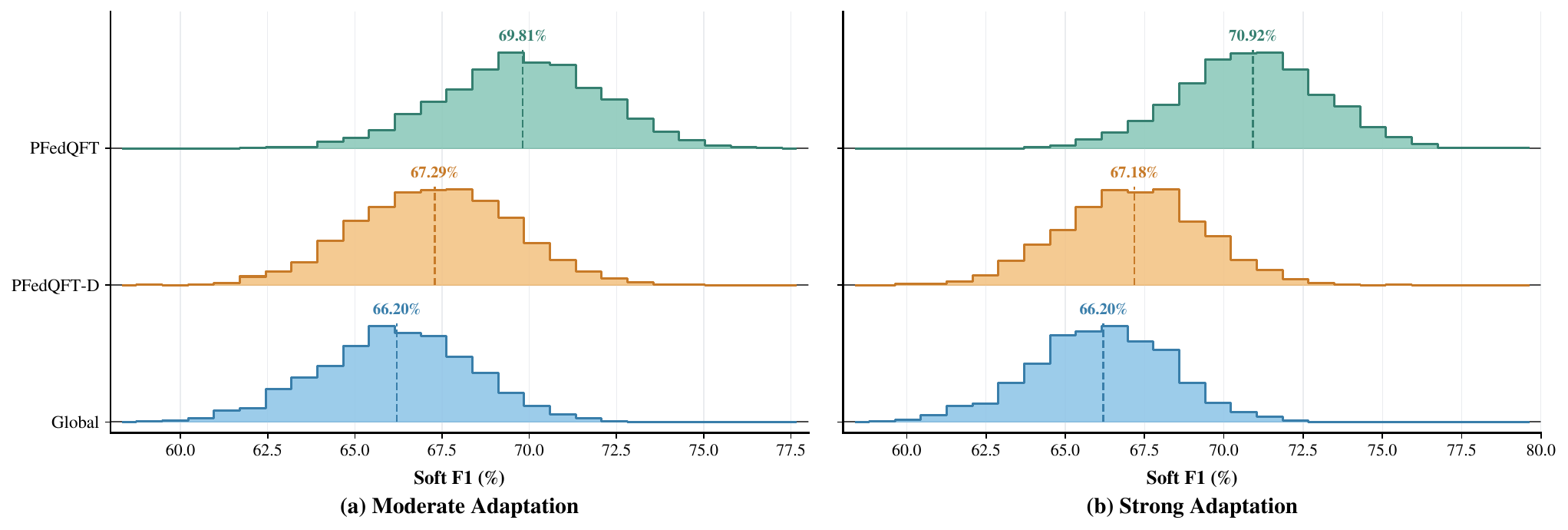}
\caption{Personalization mechanism ablation on the 196-example CHIP-CDN target-client validation set. Global disables private adaptation; \model-D trains the private branch on local examples whose diagnosis concepts do not overlap with the validation set; and \model uses local support covering the target concepts. Moderate and Strong denote private-branch update settings $(\eta_p,s_p,\lambda_p)=(0.0015,0.12,5\!\times\!10^{-5})$ and $(0.0020,0.16,2\!\times\!10^{-5})$, respectively. Shared parameters remain frozen in all personalized runs. Each distribution is obtained from 3,000 bootstrap resamples of the validation examples rather than independent training runs; dashed lines indicate Soft F1 on the complete validation set.}
\label{fig:personalization-ablation}
\end{figure*}

\subsubsection{Overall Performance Comparison}
\label{sec:overall-performance}
To assess overall effectiveness across medical language understanding tasks, we compare \model with four FFT baselines based on LoRA using Llama-3.2-3B-Instruct; accuracy is reported on the three complete MedQA validation splits, whereas Soft F1 is used for the four CBLUE tasks. Figure~\ref{fig:overall-performance} shows that \model achieves the highest accuracy on MedQA-MCMLE (81.75\%), exceeding FLoRA by 0.29 percentage points, and obtains 76.91\% and 65.25\% on MedQA-TWCMLE and MedQA-USMLE, respectively, remaining within 1.49 points of the strongest baseline. On CHIP-CDN and CMeEE-V2, \model reaches 76.71\% and 77.46\% Soft F1, only 0.33 and 0.26 points below FLoRA. On CMeIE and CMedCausal, \model obtains 59.48\% and 63.09\% Soft F1, respectively, trailing FLoRA by 5.58 and 6.10 points. Relative to the strongest baseline on each task, the corresponding gaps are 6.35 points to FedEx-LoRA and 7.73 points to FedAdam-LoRA. These losses may reflect capacity constraints specific to each task imposed by aggressive parameter compression and basis sharing across layers when modeling complex entity relations and causal links. Overall, \model ranks first or remains within 1.5 points of the best baseline on five of the seven datasets, while the CMeIE and CMedCausal results reveal a trade-off between parameter efficiency and predictive performance that depends on the task.

\subsubsection{Performance across Backbone Models}
\label{sec:backbone-comparison}
To evaluate portability across model families and parameter scales, we compare \model with FLoRA on the three complete MedQA validation splits using five general-purpose and medically adapted backbones ranging from 1.7B to 8B parameters. Across the 15 combinations in Fig.~\ref{fig:backbone-performance}, \model outperforms FLoRA in seven cases, with the most consistent gains obtained by Llama-3.1-8B-Instruct: 1.28, 0.28, and 1.18 percentage points on MedQA-MCMLE, MedQA-TWCMLE, and MedQA-USMLE, respectively; this backbone also produces the highest MedQA-TWCMLE and MedQA-USMLE accuracies among the evaluated models. On MedQA-TWCMLE, \model further improves Qwen3-1.7B and Qwen3-4B by 0.64 and 0.71 points. The two methods differ by at most 0.24 points for Llama-3.2-3B on MedQA-USMLE and for MedGemma on MedQA-TWCMLE and MedQA-USMLE, whereas \model trails FLoRA on MedQA-MCMLE by 2.89, 2.04, and 7.53 points with Qwen3-1.7B, Qwen3-4B, and MedGemma. These results confirm broad applicability but also indicate that the attainable gains depend on the backbone, with Llama-3.1-8B-Instruct providing the most stable improvements.

\subsubsection{Sensitivity to Data Heterogeneity}
\label{sec:heterogeneity-sensitivity}
To assess sensitivity to heterogeneous client data, we evaluate FLoRA and \model under three Dirichlet concentrations $\beta\in\{1.0,0.5,0.1\}$ using five clients, with matched client assignments and identical global validation examples. Table~\ref{tab:heterogeneity} reports the performance of the global adapter after the final round with private residuals cleared. Across the three concentrations, the performance ranges of \model are only 0.35, 0.49, and 0.39 percentage points on MedQA-MCMLE, CHIP-CDN, and CMeEE-V2, respectively, compared with 0.68, 1.99, and 0.67 for FLoRA. The reduction is most pronounced on CHIP-CDN, where the range of \model is approximately four times smaller. Under severe heterogeneity, \model also achieves 81.14\% accuracy on MedQA-MCMLE, closely matching the 81.15\% of FLoRA. These consistently narrower variations demonstrate that \model maintains more stable performance across the evaluated heterogeneity levels and is less sensitive to changes in client data distributions.

\subsubsection{Parameter and Communication Efficiency}
\label{sec:parameter-communication-efficiency}
Resource efficiency is evaluated by averaging end-to-end runtime, trainable parameters, adapter storage, and upload volume for each client across the seven tasks. As reported in Table~\ref{tab:efficiency}, \model uses an average of 1.6524 million trainable parameters, or 0.0514\% of the backbone, and transmits 6.5981 MB per client per round; these values represent reductions of 93.20\% and 93.22\%, respectively, relative to FLoRA. Its residuals retained privately by each client occupy only 0.011648 MB on average and require no transmission. The average end-to-end runtime is 6 h 23 min under the evaluated task-specific configurations, reflecting the additional structured basis construction and personalized residual optimization. Overall, \model provides a compact parameter and communication footprint across the seven medical tasks.

\subsubsection{Ablation of the Personalization Mechanism}
\label{sec:personalization-ablation}
The contribution of client private adaptation is evaluated on CHIP-CDN using a target client validation set of 196 examples. All conditions start from the same global checkpoint and retain the same model structure; during eight additional rounds, the shared parameters remain frozen and only the private residual coefficients are optimized. Global disables the private residual operator and evaluates the federated global adapter directly. \model-D trains the private branch using local examples whose diagnosis concepts do not overlap with those in the target validation set, whereas \model uses a local support set that covers the target concepts. Moderate and Strong refer to the update strength of the private branch rather than the degree of data heterogeneity: they use $(\eta_p,s_p,\lambda_p)=(0.0015,0.12,5\!\times\!10^{-5})$ and $(0.0020,0.16,2\!\times\!10^{-5})$, respectively, while keeping the adaptation rank, intrinsic rank, Taylor order, and effective batch size fixed. The plotted distributions are obtained from 3,000 bootstrap resamples of the same validation predictions and therefore characterize evaluation variability rather than variation across independent training runs.

As shown in Fig.~\ref{fig:personalization-ablation}, Global achieves 66.20\% Soft F1 under both configurations. With Moderate adaptation, \model-D and \model obtain 67.29\% and 69.81\%, respectively; with Strong adaptation, their scores are 67.18\% and 70.92\%. Thus, \model improves over Global by 3.61 and 4.71 percentage points and over \model-D by 2.53 and 3.74 points under the Moderate and Strong settings, respectively. \model-D yields an improvement of only approximately one percentage point and does not benefit from the stronger private update, whereas \model gains a further 1.11 points. The rightward shift of the \model bootstrap distributions demonstrates that the personalization gain is primarily associated with support data relevant to the target client, and that increasing private adaptation strength is effective when the local concepts align with the target distribution.

\section{Conclusion}
\label{sec:conclusion}
This work presents \model, a framework inspired by quantum transformations for efficient personalized FFT of medical language models. By combining shared structured adaptation subspaces, full core transformations for individual projections, and structured residual operators controlled by locally retained coefficients, the framework supports collaborative knowledge transfer and adaptation to the local data distribution of each participating institution. Experiments on medical datasets demonstrate substantial reductions in trainable parameters and communication costs while maintaining competitive predictive performance. Several limitations warrant further investigation. Broader validation across additional medical domains, larger backbone models, and deployments involving multiple institutions is needed to establish the robustness and generalizability of these findings. Retaining institution specific parameters locally does not itself provide formal privacy guarantees, and the clinical reliability of model outputs remains unverified. Future work will explore aggregation with formal privacy guarantees, further reductions in communication costs, and more effective client personalization mechanisms. Evaluation in clinical settings will also be necessary to assess reliability and suitability for medical applications before deployment.

\bibliographystyle{IEEEtran}
\bibliography{references}

\end{document}